\documentclass[10pt]{article}
\usepackage{bitinf}

\reportid{BitInf-TR-2026-06}
\reportdate{June 2026}
\papertitle{Arko-T: A Foundation Model for Text-to-Structured 3D Generation}
\runningtitle{Arko-T}
\paperauthors{%
Liang Wang\textsuperscript{*\ddag\,1,2}\quad
Zhaoyang Xi\textsuperscript{*\ddag\,1,2}\quad
Zekai Xiang\textsuperscript{*\ddag\,1,3}\quad
Heng Meng\textsuperscript{*\,1,2}\quad
Qishan Zhang\textsuperscript{*\,1,2}\\[2pt]
Pingyi Zhou\textsuperscript{\dag\,1}\quad
Jin Liu\textsuperscript{\dag\,2}\quad
Litao Chen\textsuperscript{1}%
}
\paperaffil{%
\textsuperscript{1}\,Spatial Design Intelligence Lab, BitInf Ltd., Shanghai 200003, China\\
\textsuperscript{2}\,School of Computer Science, Wuhan University, Wuhan 430000, Hubei, China\\
\textsuperscript{3}\,College of Computer and Information Engineering, Nanjing Tech University, Nanjing 211800, Jiangsu, China\\[4pt]
\textsuperscript{*}\,Work done during internship at Spatial Design Intelligence Lab, BitInf Ltd.\quad
\textsuperscript{\ddag}\,Equal contribution.\quad
\textsuperscript{\dag}\,Corresponding authors.%
}
\papercontact{%
Correspondence: \href{mailto:zhoupingyi@bitinf.ai}{zhoupingyi@bitinf.ai}, \href{mailto:jinliu@whu.edu.cn}{jinliu@whu.edu.cn}%
}

\newcommand{\arkot}{Arko-T}

\usepackage{pifont}
\usepackage{float}
\usepackage{placeins}
\usepackage{listings}
\definecolor{codekw}{HTML}{1E2761}
\definecolor{codecomment}{HTML}{6B7280}
\definecolor{codestr}{HTML}{0F766E}
\begin{document}

\makebitinftitle

\begin{figure}[!h]
  \centering
  \includegraphics[width=0.8\linewidth]{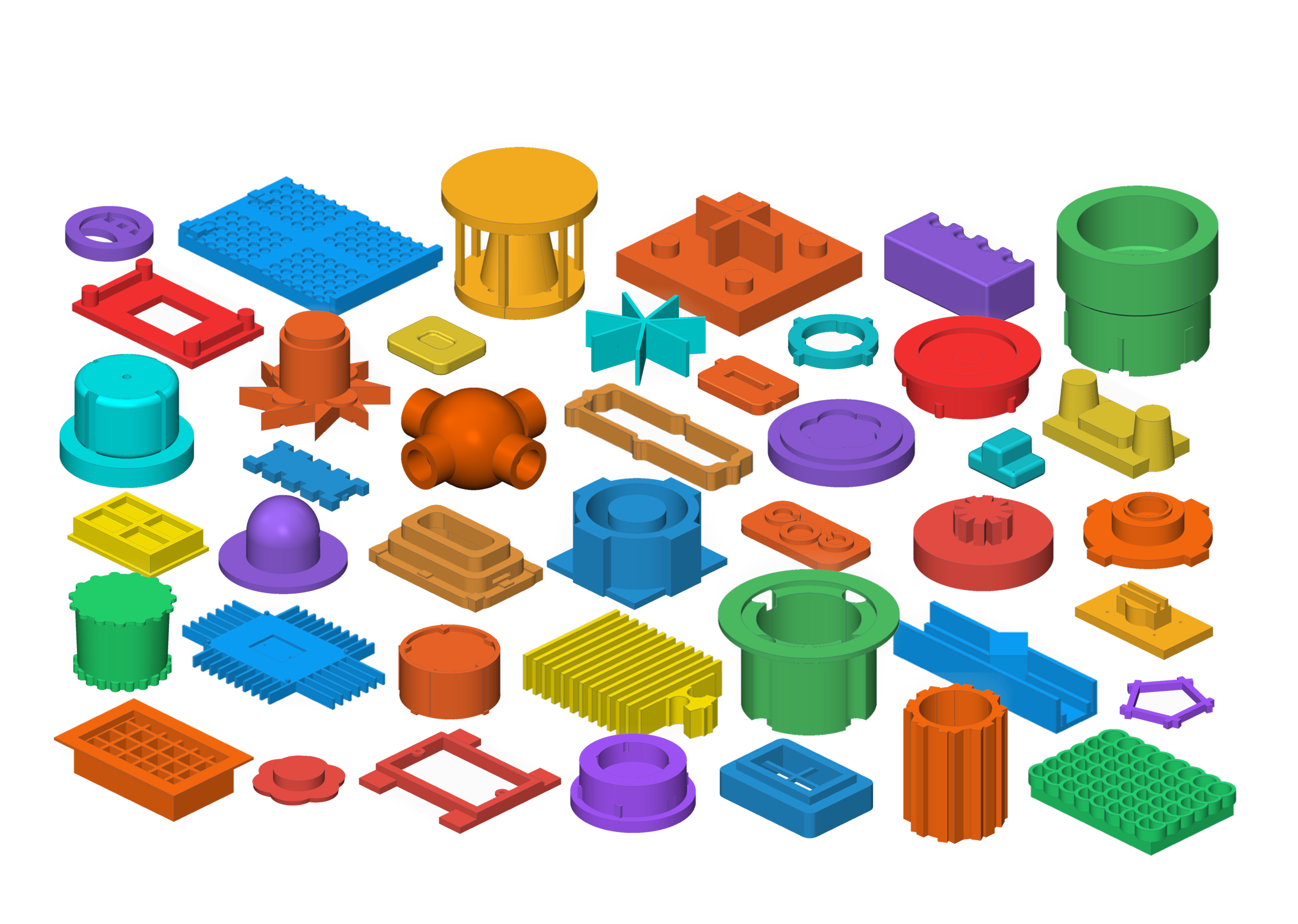}
  \caption{Out-of-distribution designs generated by \arkot{}. None of these parts appears in the training data; the gallery illustrates spatial reasoning and the ability to compose mechanical features into coherent structured designs.}
  \label{fig:hero}
\end{figure}
\vspace*{\fill}
\clearpage

\begin{abstract}
Text-to-3D systems can now synthesize a model from a single sentence, yet the result is a shape to render, not a design to edit. We present \textbf{\arkot{}}, a $4$B-parameter text-to-design model that maps natural-language intent directly into executable, parametric CAD programs. Rather than optimizing for code executability alone, \arkot{} aligns every stage of the pipeline to a formal notion of \emph{design state}, so that data curation, code normalization, and execution-grounded supervision all work to preserve the features, parameters, and construction logic that make a CAD artifact editable. Benchmarked against seven frontier LLMs across $12$ metrics, \arkot{} attains the best score on $8$ and the second-best on $3$ more, at roughly one-tenth the per-benchmark cost. The results suggest that targeted design-level training at moderate scale can match frontier general-purpose models on structured CAD generation.
\end{abstract}

\section{Introduction}
\label{sec:intro}

Text-to-3D systems can now generate a photorealistic bracket from a single sentence, but no engineer can change the hole spacing in the result. The output is a \emph{shape}, not a \emph{design}. A design is a structured artifact: it carries named features (holes, ribs, fillets), adjustable parameters (thickness, spacing, radius), a construction history that records how the part was built, and references that bind features to faces and sketches. Producing that artifact from natural language, what we call \emph{text-to-design}, is the problem this report addresses.

\begin{figure}[!h]
  \centering
  \includegraphics[width=0.6\linewidth]{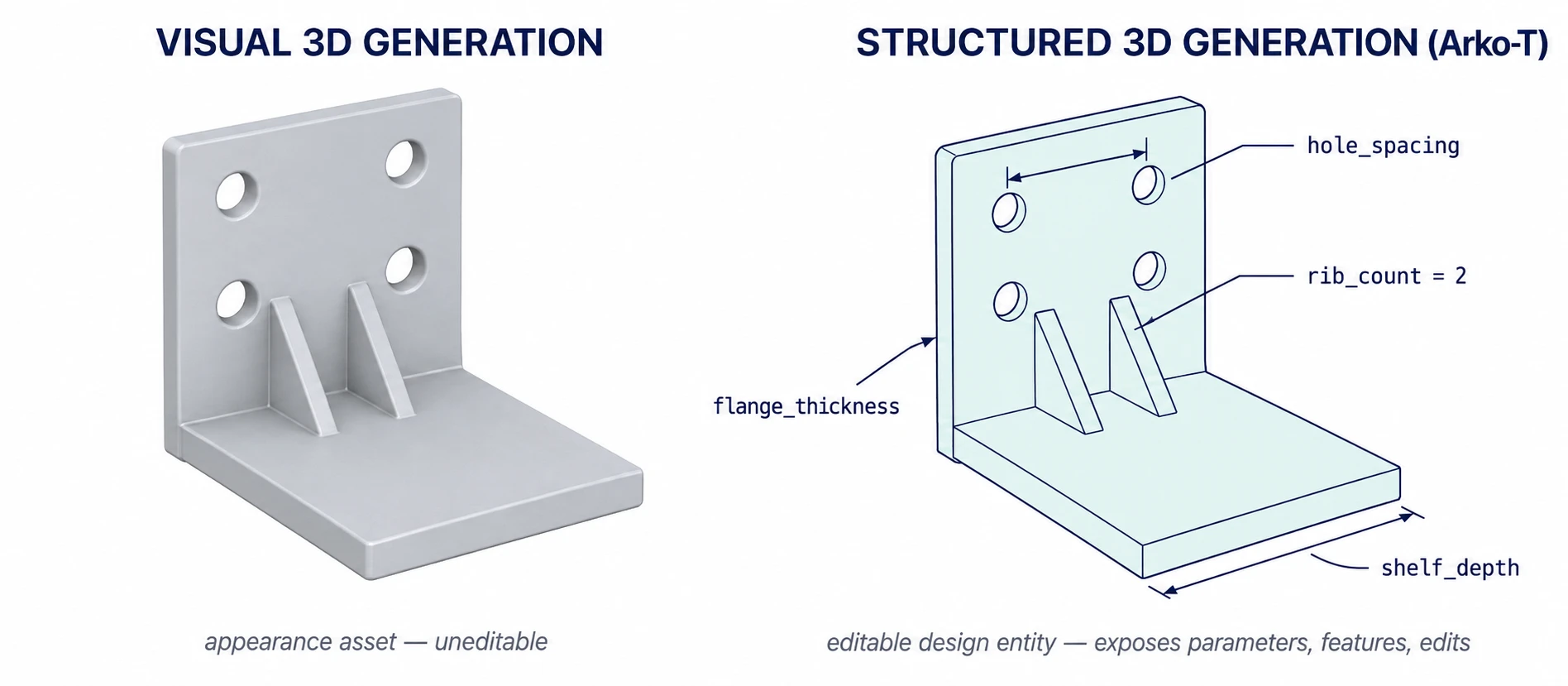}
  \caption{Visual 3D asset vs.\ structured design entity. A visual mesh (left) renders correctly but offers no editable parameters. A parametric CAD program (right) exposes named dimensions, feature structure, and construction history.}
  \label{fig:teaser}
\end{figure}

Existing methods have moved steadily toward this goal without reaching it. Visual text-to-3D systems~\citep{poole2022dreamfusion,lin2022magic3d,cheng2023sdfusion} synthesize meshes and neural fields that render correctly but carry no parametric structure; changing a dimension means regenerating from scratch (\cref{fig:teaser}). Text-to-CAD methods~\citep{khan2024text2cad,li2025cadllama,guan2025cadcoder,pyatov2026cadfs,niu2025cadrl,li2025recad} take a further step by producing executable CAD code, but their focus has leaned toward making programs \emph{runnable} rather than ensuring they realize the design intent a prompt describes; outside their training distribution, these models tend to fall back on basic geometric primitives. The progression from visual fidelity to code executability is real, but neither station delivers what an engineer actually needs: an editable, feature-bearing design entity.

We present \arkot{}, a text-to-design model that aligns every stage of the pipeline to a formal notion of \emph{design state}. We first formalize this design state and the criteria under which a generated program counts as a useful design (\cref{sec:problem}); then build a training corpus of $1.3$M Build123d programs whose code is normalized to expose that structure, and train a $4$B-parameter model through execution-grounded supervision so that the learning signal is a valid \emph{design}, not merely a valid \emph{program} (\cref{sec:arkot}); finally, evaluate whether the resulting outputs behave as designs, executable, geometrically faithful, and feature-bearing, rather than only as code that compiles (\cref{sec:experiments}).

This report makes three contributions.
\begin{enumerate}[leftmargin=*,topsep=2pt,itemsep=1pt]
  \item We formalize \emph{text-to-design} as a task distinct from text-to-CAD code generation, defining the design state a useful output must preserve and an evaluation framework that measures executability, geometric agreement, and feature realization (\cref{sec:problem,sec:output-criteria}).
  \item We describe the data pipeline, code normalization, and two-stage training procedure that connect design-state formalization to a trainable model (\cref{sec:arkot}).
  \item We evaluate \arkot{} against seven frontier LLMs on $12$ quantitative metrics. At roughly $10$--$60{\times}$ lower per-benchmark cost, \arkot{} achieves the top score on $8$ metrics and second-best on $3$ more (\cref{sec:main-results}).
\end{enumerate}

\section{Related Work}
\label{sec:related}

Three lines of work address different aspects of generating 3D artifacts from language. Each advances the field in one direction and leaves a specific gap.

\paragraph{Visual 3D from language.}
DreamFusion, Magic3D, SDFusion, and their successors~\citep{poole2022dreamfusion,lin2022magic3d,cheng2023sdfusion} established that natural language is a viable conditioning signal for 3D output. Their outputs, however, are appearance assets---meshes or neural fields optimized for visual plausibility---and carry no parametric structure. Changing a hole diameter or adding a rib requires re-generation from scratch.

\paragraph{Learning CAD structure.}
A parallel line treats CAD geometry and construction processes as learnable objects. Dataset efforts such as ABC~\citep{koch2019abc}, Fusion 360 Gallery~\citep{willis2021fusion360}, SketchGraphs~\citep{seff2020sketchgraphs}, and DeepCAD~\citep{wu2021deepcad} progressively move from static geometry toward construction histories and constraint graphs. Recent work extends this to design constraints as first-class objects~\citep{casey2025aligning,lin2026unisketch}, direct B-rep generation~\citep{xu2024brepgen,guo2025brepgiff,li2025dtgbrepgen,li2025caddreamer}, and reverse engineering of executable code from point clouds~\citep{rukhovich2024cadrecode}. These results show that CAD structure is learnable, but most operate without natural-language input.

\paragraph{Text-to-CAD as code generation.}
Text2CAD~\citep{khan2024text2cad}, CAD-Llama~\citep{li2025cadllama}, Text-to-CadQuery~\citep{xie2025textcadquery}, CAD-Coder~\citep{guan2025cadcoder}, CADFS~\citep{pyatov2026cadfs}, CAD-RL~\citep{niu2025cadrl}, and ReCAD~\citep{li2025recad} connect language prompts to executable CAD programs. This line demonstrates feasibility, but the emphasis has leaned toward \emph{executability}---whether the code runs---over \emph{design fidelity}---whether the output preserves the features, parameters, and construction intent the prompt describes. In practice, these models perform well within their training distribution yet degrade sharply on out-of-distribution prompts, often reverting to simple primitives rather than composing the mechanical features the request names.

\arkot{} draws on all three threads---language conditioning, CAD-native data, and executable code---but shifts the target from code that compiles to \emph{design states} that a human designer can inspect and edit.

\section{Problem Definition}
\label{sec:problem}

We define text-to-design as the task of mapping a natural-language design request~$x$ into a structured \emph{design state}~$z$ that can be decoded into a backend-specific CAD program, executed, and inspected:
\begin{equation}
  x \;\rightarrow\; z \;\xrightarrow{D_b}\; p_b \;\xrightarrow{E_b}\; (g,\ell),
  \label{eq:task-chain}
\end{equation}
where $D_b$ is a backend decoder, $p_b$ the emitted program, $E_b$ the CAD execution environment, $g$ the resulting geometry, and $\ell$ the execution log. In the current system $b$~is Build123d; the formalization is backend-agnostic.

The design state captures five components:
\begin{equation}
  z = (\mathcal{F},\;\Theta,\;\mathcal{C},\;\mathcal{H},\;\mathcal{A}),
  \label{eq:design-state}
\end{equation}
where $\mathcal{F}$ is a \emph{feature vocabulary} (holes, ribs, fillets, shells, patterns, \ldots), $\Theta$ a set of \emph{named parameters} (radii, thicknesses, spacings), $\mathcal{C}$ \emph{constraints and relations} (symmetry, coplanarity, spacing rules), $\mathcal{H}$ a \emph{construction history} (the ordered sequence of sketch, extrude, cut, fillet, \ldots\ operations), and $\mathcal{A}$ \emph{attachments} (references binding features to faces, edges, or sketch planes). For a bracket with four mounting holes and triangular ribs, $\mathcal{F}$ includes \texttt{hole} and \texttt{rib}; $\Theta$ includes \texttt{hole\_radius} and \texttt{plate\_thickness}; $\mathcal{C}$ includes the spacing relation among holes; $\mathcal{H}$ records the sketch--extrude--cut--fillet order; and $\mathcal{A}$ binds each hole to a specific face.

This formalization serves a concrete role in our pipeline: it defines what the training data must expose (\cref{sec:data}) and what the code normalization must preserve (\cref{sec:normalization}). \Cref{tab:representation-contrast} compares common 3D representations against these five properties.

\begin{table}[!ht]
  \caption{3D representations compared on design-state properties. }
  \label{tab:representation-contrast}
  \centering
  \small
  \setlength{\tabcolsep}{5pt}
  \newcommand{\Yes}{\ding{51}}
  \newcommand{\No}{\ding{55}}
  \begin{tabular}{@{}lccccc@{}}
    \toprule
    Representation & Render & Param.\ edit & Design intent & Feature struct.\ & CAD workflow \\
    \midrule
    Mesh                    & \Yes & \No  & \No  & \No  & \No  \\
    Point cloud             & \Yes & \No  & \No  & \No  & \No  \\
    NeRF / 3DGS             & \Yes & \No  & \No  & \No  & \No  \\
    B-rep                   & \Yes & \No  & \No  & \Yes & \Yes \\
    Parametric CAD program  & \Yes & \Yes & \Yes & \Yes & \Yes \\
    \bottomrule
  \end{tabular}
\end{table}

\paragraph{Distinction from code generation.}
A syntactically valid CAD script can produce an empty body, a non-manifold solid, or a shape that bears no relation to the requested features. Text-to-design requires more than compilation: the output must be a valid solid \emph{and} preserve the design state that the prompt implies.

\paragraph{Distinction from B-rep generation.}
A B-rep captures exact boundary topology but does not, by itself, preserve the feature history, named parameters, or construction intent behind that geometry (\cref{tab:representation-contrast}). Text-to-design targets the full design state, not only the boundary of the executed solid.

\section{Arko-T}
\label{sec:arkot}

\arkot{} instantiates the design-state formalization of \cref{sec:problem} as a trainable system. This section describes the model, the data it learns from, how that data is normalized to expose design-state structure, and the training procedure.

\begin{figure*}[!b]
  \centering
  \includegraphics[width=\linewidth]{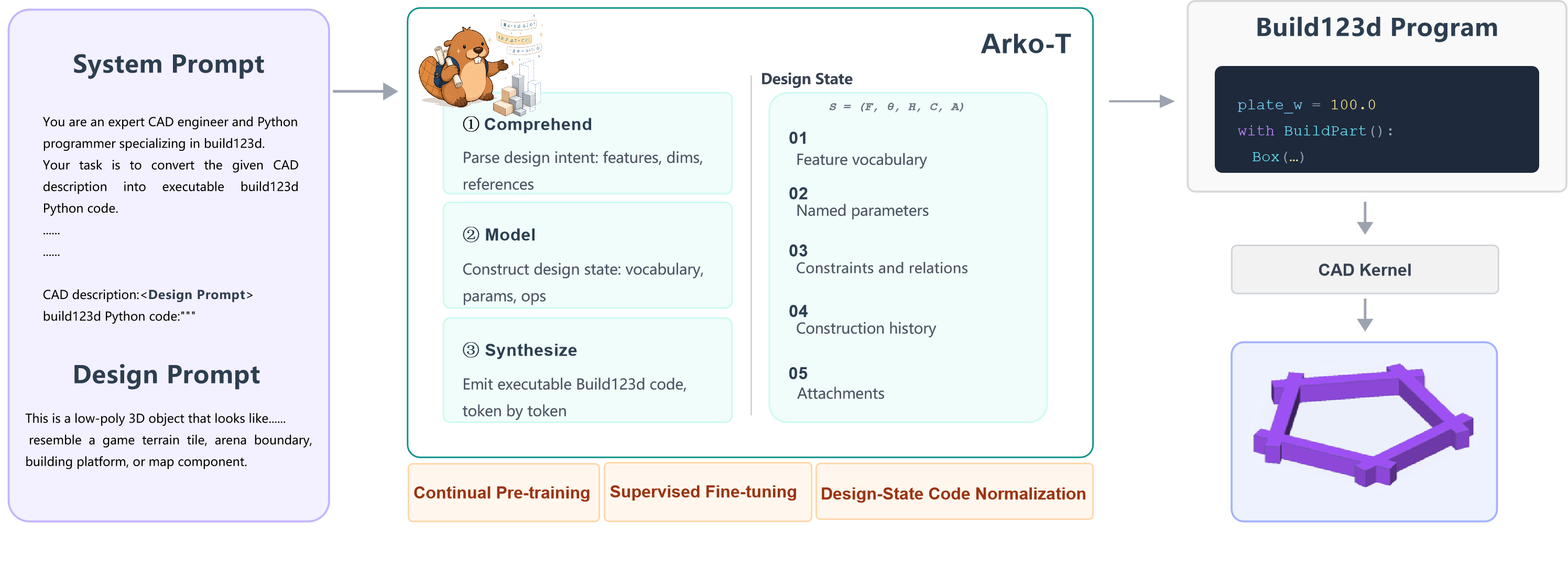}
  \caption{\arkot{} pipeline. A natural-language design prompt enters the model; an executable Build123d program is emitted; running the program through the CAD kernel yields an editable, parameterized design entity.}
  \label{fig:architecture}
\end{figure*}

\subsection{Model}
\label{sec:model}

\arkot{} is a $4$B-parameter transformer initialized from Qwen3.5-4B~\citep{qwen2024qwen25} and adapted to parametric CAD through two training stages described in \cref{sec:training}. The model operates as a sequence-to-sequence generator: the input is a single-part text prompt, and the output is a Build123d~\citep{build123dDocs} program that defines named parameters, constructs geometry, and composes the full vocabulary of parametric CAD operations into a solid object (\cref{fig:architecture}).

\subsection{Training Data}
\label{sec:data}

The training corpus comprises approximately $1.3$M (prompt, Build123d program) pairs drawn from two sources in roughly equal proportion.

\paragraph{Dataset conversion.}
Existing CAD datasets---DeepCAD~\citep{wu2021deepcad} and related collections---provide large-scale construction sequences in their native formats. We convert these sequences into executable Build123d programs and pair them with generated text descriptions, discarding any program that fails CAD-kernel execution.

\paragraph{Open-source and internal curation.}
A second source draws on open-source CAD repositories and internally authored designs. Programs are cleaned, standardized, and paired with natural-language descriptions. This source broadens domain coverage beyond the industrial parts that dominate existing datasets to include consumer products, architectural components, and personal-fabrication designs.

Both sources pass through execution filtering: every program is run through the Build123d kernel, and only programs that produce a valid, non-empty CAD solid enter the training set. This ensures that the supervision signal is grounded in geometric validity, not surface plausibility.

\subsection{Design-State Code Normalization}
\label{sec:normalization}

Raw CAD programs vary widely in style: variable names may be opaque, construction steps may be interleaved with unrelated logic, and design parameters may be buried as inline constants. Such variation teaches the model incidental formatting rather than design structure.

\begin{figure}[t]
  \centering
  \begin{minipage}[c]{0.62\linewidth}
\begin{lstlisting}
# Named design parameters (mm)
plate_width       = 100.0
vent_slot_count   = 5
vent_slot_spacing = 15.0
rib_width         = 12.0
...

# Through-cut ventilation slot
Box(vent_slot_width, vent_slot_height,
    plate_thickness + 2.0,
    mode=Mode.SUBTRACT)
...

# Louver rib, linearly patterned across the slots
for i in range(vent_slot_count):
    add(rib.moved(Location((0.0,
        i * vent_slot_spacing, 0.0))))
\end{lstlisting}
  \end{minipage}\hfill
  \begin{minipage}[c]{0.34\linewidth}
    \centering
    \includegraphics[width=\linewidth]{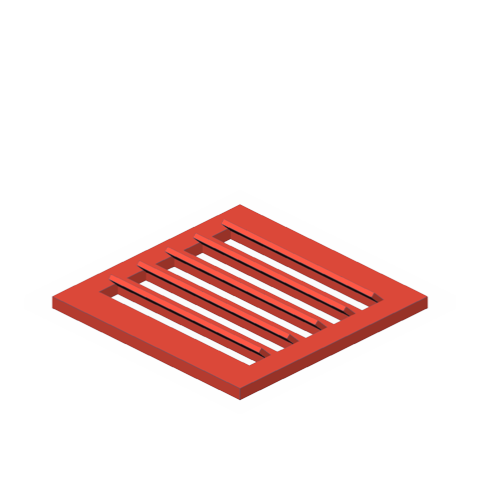}
  \end{minipage}
\caption{Representative \arkot{} output (abridged). Design dimensions are surfaced as named parameters; construction follows a consistent feature--operation order.}
\label{fig:param-listing}
\end{figure}

We normalize every training program to expose the five components of the design state (\cref{eq:design-state}). For \textbf{feature vocabulary}~$(\mathcal{F})$, each mechanical feature (hole, rib, fillet, shell, pattern, \ldots) is realized through a canonical code pattern, so the model learns to associate feature names with construction idioms. For \textbf{named parameters}~$(\Theta)$, dimensions that a designer would want to adjust---radii, thicknesses, spacings, counts---are extracted to a parameter block at the top of the program with descriptive names and unit annotations (\cref{fig:param-listing}). For \textbf{construction history}~$(\mathcal{H})$, operations follow a consistent order (sketch $\to$ extrude $\to$ secondary features $\to$ finishing) so that the model acquires reusable construction sequences rather than arbitrary orderings. For \textbf{constraints and references}~$(\mathcal{C},\mathcal{A})$, spatial relations (symmetry, spacing) and face/edge references are expressed explicitly rather than computed implicitly, making the design logic inspectable.

The result is a training corpus in which every program is simultaneously executable code \emph{and} a readable specification of design intent.

\subsection{Training Procedure}
\label{sec:training}

Training proceeds in two stages.

\paragraph{Stage~1: Continual pre-training.}
The base Qwen3.5-4B model is pre-trained on a corpus of 3D-design-related text---documentation and API references from multiple parametric CAD tools (Build123d, CadQuery, OpenSCAD, and others), open-source modeling libraries, and internal design documentation---to build domain vocabulary and acquaint the model with the language of parametric construction before it sees paired data.

\paragraph{Stage~2: Supervised fine-tuning.}
The pre-trained checkpoint is fine-tuned on the $1.3$M (prompt, program) pairs using LoRA~\citep{hu2022lora}. Training runs for one epoch with a batch size of $256$ and a learning rate of $2 \times 10^{-4}$ on $16$~GPUs. Because every training program has already passed execution filtering (\cref{sec:data}), the supervised signal is a design-valid program, not merely a syntactically correct one.

\subsection{Feature and Domain Coverage}
\label{sec:coverage}

The training corpus is curated to cover two axes simultaneously. Along the \emph{feature axis}, the data spans the full range of mechanical features and parametric operations that engineering parts require---from basic extrusions and cuts through advanced features such as shells, sweeps, and polar arrays. Along the \emph{domain axis}, it includes industrial parts (brackets, housings, flanges), consumer products, architectural elements, and personal-fabrication designs. This dual coverage is what enables \arkot{} to compose features into coherent designs for out-of-distribution prompts (\cref{fig:hero}), rather than falling back on basic primitives when the request leaves the training distribution.

\section{Experiments}
\label{sec:experiments}

\subsection{Benchmark and Metrics}
\label{sec:output-criteria}
\label{sec:benchmark}

We evaluate on the Text2CAD-Bench benchmark~\citep{wang2026text2cad}, which comprises $400{+}$ CAD prompts spanning single-feature parts (L1) and multi-feature parts (L2). Each prompt has two stylistic variants that mirror the two registers a CAD assistant encounters in practice. \textbf{Geo}~(geometric) prompts are lay-user descriptions that name dimensions and shapes as a non-expert would phrase a request; \textbf{Pro}~(procedural) prompts are domain-expert construction sequences that specify operations in the order a CAD professional would author them.

A generated program must pass two tests before it can be scored: it must execute without error, and the result must be a non-empty, valid CAD solid. We report three quantitative metrics per split.

\textbf{Invalid rate~(IR)} is the fraction of prompts that fail this validity gate---the program either crashes or produces an empty or non-manifold body (lower is better). 

\textbf{Chamfer Distance~(CD)} measures surface discrepancy between the generated and ground-truth shapes:
\begin{equation}
  \mathrm{CD}(P,Q) = \frac{1}{|P|}\sum_{p \in P}\min_{q \in Q}\lVert p-q\rVert^{2}
  + \frac{1}{|Q|}\sum_{q \in Q}\min_{p \in P}\lVert q-p\rVert^{2},
  \label{eq:cd}
\end{equation}
reported $\times 10^{3}$ (lower is better). CD is sensitive to local deviations: a missing hole or an extra rib produces a measurable surface gap even if the overall silhouette is correct. 

\textbf{Intersection over Union~(IoU)} measures global shape overlap:
\begin{equation}
  \mathrm{IoU} = \frac{|V_1 \cap V_2|}{|V_1 \cup V_2|},
  \label{eq:iou}
\end{equation}
where $V_1, V_2$ are occupied voxel grids at $128^{3}$ resolution (higher is better). IoU captures overall volumetric agreement but is less sensitive to fine-grained feature differences. Before scoring, both shapes are aligned in scale and pose through bidirectional normalization. CD and IoU are computed only on programs that pass the validity gate. Because both metrics can still score well when individual features are absent, we supplement them with qualitative analysis of feature realization (\cref{sec:qualitative}).

\subsection{Baselines and Protocol}
\label{sec:baselines}

We compare \arkot{} against seven frontier general-purpose LLMs: Gemini-3.5-flash~\citep{gemini2024gemini15}, DeepSeek-V4-Pro~\citep{deepseek2024v3}, GPT-5.2~\citep{openai2023gpt4}, Qwen3.6-max~\citep{qwen2024qwen25}, Kimi-k2.6~\citep{kimi2025k15}, Claude-4.5-sonnet~\citep{anthropic2024claude3}, and GLM-5.1~\citep{glm2024chatglm}. Each baseline is the vendor's single best-performing model at the time of evaluation. All models receive the same prompt, are sampled at temperature $0.6$, and are allowed up to $3$ retry attempts when the generated program fails CAD-kernel execution. The ``Requests'' column in \cref{tab:efficiency} reflects total calls including retries.

\paragraph{Why general-purpose LLMs rather than specialized text-to-CAD models?}
Existing specialized models~\citep{khan2024text2cad,li2025cadllama,guan2025cadcoder} are trained and evaluated within their own dataset distributions. When prompted with the design-level requests in our benchmark---which name specific mechanical features, spatial relations, and construction intent beyond those distributions---these models degrade sharply, falling back on basic geometric primitives rather than producing feature-rich designs. They address a different task---text-to-CAD-code within a fixed distribution---and are not designed for open-ended text-to-design. We therefore benchmark against frontier LLMs, which can in principle handle arbitrary design intent through their general code-generation capability and represent the strongest available alternative.

\subsection{Main Results}
\label{sec:main-results}

\Cref{tab:main-results} summarizes the results. On Chamfer Distance, \arkot{} ranks first on all four splits ($2.46$, $2.33$, $4.92$, $5.02$), indicating that its generated shapes are consistently closer to the ground truth than those of any baseline. On IoU, \arkot{} leads on both L2 splits (Geo~$0.801$, Pro~$0.780$) and is the runner-up on L1~Geo ($0.873$ vs.\ Gemini's $0.883$). On invalid rate, \arkot{} leads on L1 ($4.40$\%, $3.90$\%) and is the runner-up on L2, where Gemini-3.5-flash achieves $1.60$\% and $9.50$\%.

\begin{figure*}[!ht]
  \centering
  \includegraphics[width=\linewidth]{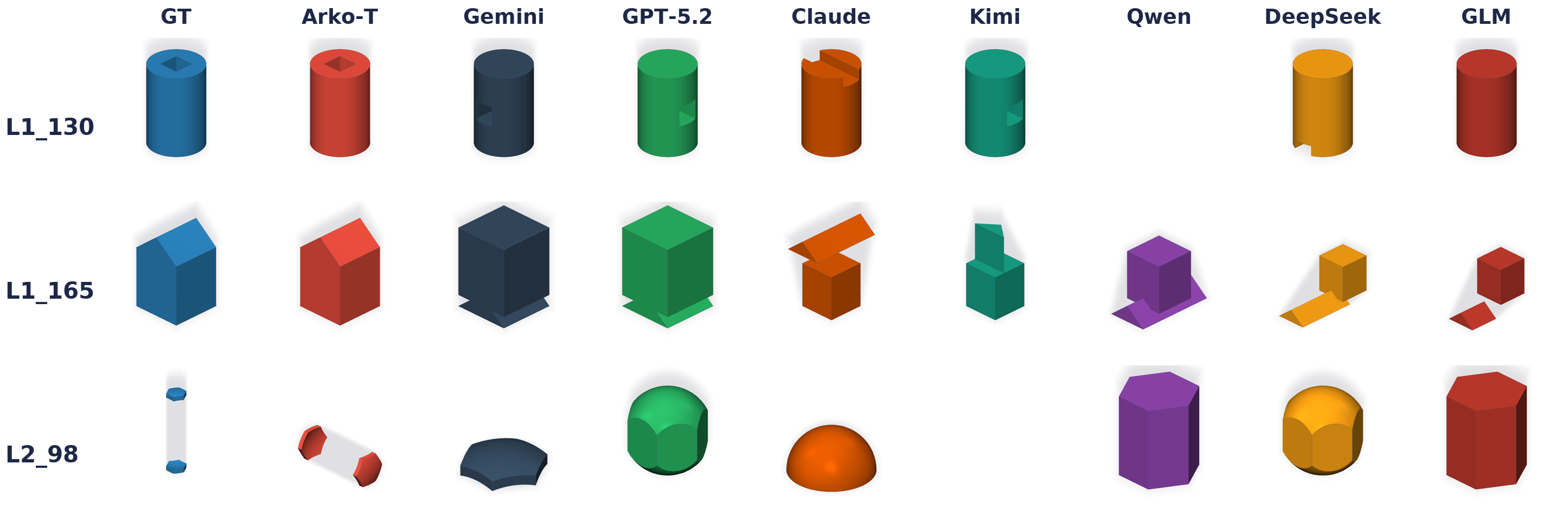}
  \caption{Qualitative comparison on three benchmark items across all models. \arkot{} produces designs that more closely match the target geometry and preserve mechanical features.}
  \label{fig:results-grid}
\end{figure*}

The one column where \arkot{} falls below second place is L1~Pro IoU ($0.868$, third behind Gemini and DeepSeek-V4-Pro). The remaining gap is concentrated on L2 invalid rate, where Gemini-3.5-flash holds a ${\sim}10$-percentage-point advantage---a gap that points to the remaining challenge of robust multi-feature construction. \Cref{fig:results-grid} visualizes three representative benchmark items across all models; \arkot{} produces designs that more faithfully realize the requested features and spatial structure.

\begin{table*}[!ht]
  \caption{Text-to-CAD results. \textbf{Bold}: best; \underline{underline}: second-best per column.}
  \label{tab:main-results}
  \centering
  \footnotesize
  \setlength{\tabcolsep}{5pt}
  \begin{tabular}{@{}l ccc ccc ccc ccc@{}}
    \toprule
    & \multicolumn{3}{c}{L1 Geo} & \multicolumn{3}{c}{L1 Pro} & \multicolumn{3}{c}{L2 Geo} & \multicolumn{3}{c}{L2 Pro} \\
    \cmidrule(lr){2-4}\cmidrule(lr){5-7}\cmidrule(lr){8-10}\cmidrule(lr){11-13}
    Model & IR$\downarrow$ & CD$\downarrow$ & IoU$\uparrow$ & IR$\downarrow$ & CD$\downarrow$ & IoU$\uparrow$ & IR$\downarrow$ & CD$\downarrow$ & IoU$\uparrow$ & IR$\downarrow$ & CD$\downarrow$ & IoU$\uparrow$ \\
    \midrule
    Gemini-3.5-flash      & 7.90  & \underline{3.54}  & \textbf{0.883}    & \underline{10.80} & \underline{3.20}  & \textbf{0.889}    & \textbf{1.60}    & 8.61              & \underline{0.783} & \textbf{9.50}    & \underline{7.98}  & \underline{0.777} \\
    \arkot{} (ours)       & \textbf{4.40}     & \textbf{2.46}     & \underline{0.873} & \textbf{3.90}     & \textbf{2.33}     & 0.868             & \underline{12.20}& \textbf{4.92}     & \textbf{0.801}    & \underline{12.20}& \textbf{5.02}     & \textbf{0.780}    \\
    DeepSeek-V4-Pro       & 11.80 & 6.96  & 0.827             & 20.20 & 5.07  & \underline{0.873} & 25.40 & 8.99  & 0.782             & 39.70 & 13.11 & 0.744             \\
    GPT-5.2               & 6.76  & 6.02  & 0.804             & 19.81 & 9.01  & 0.829             & 24.88 & \underline{8.46}  & 0.744             & 37.31 & 14.40 & 0.708             \\
    Qwen3.6-max           & 12.80 & 8.81  & 0.797             & 20.20 & 5.78  & 0.845             & 24.30 & 13.42 & 0.727             & 38.60 & 11.45 & 0.710             \\
    Kimi-k2.6             & 4.90  & 9.30  & 0.751             & 16.70 & 8.37  & 0.803             & 15.30 & 14.94 & 0.677             & 33.30 & 15.74 & 0.686             \\
    Claude-4.5-sonnet     & 14.30 & 7.69  & 0.775             & 23.20 & 8.27  & 0.795             & 41.30 & 16.19 & 0.669             & 55.00 & 15.20 & 0.658             \\
    GLM-5.1               & 15.80 & 12.86 & 0.725             & 27.10 & 9.86  & 0.792             & 27.50 & 16.82 & 0.640             & 55.00 & 11.24 & 0.717             \\
    \bottomrule
  \end{tabular}
\end{table*}

\subsection{Inference Efficiency and Cost}
\label{sec:efficiency}

\Cref{tab:efficiency} reports latency and cost. \arkot{} runs at $0.41$\,s per item on local GPUs, faster than every API baseline by at least an order of magnitude. On dollar cost, the full benchmark run costs \$$0.28$---roughly $6{\times}$ less than the cheapest API baseline (DeepSeek at \$$1.69$) and $65{\times}$ less than the most expensive (Claude at \$$18.14$). The comparison is not strictly apples-to-apples: API prices include vendor margins and infrastructure, while the \arkot{} figure reflects raw compute at \$$1.50$/A100-hour. Nevertheless, the gap is large enough to indicate a qualitative difference in deployment cost between a $4$B specialized model and frontier-scale general-purpose APIs.

\begin{table}[!ht]
  \caption{Per-benchmark inference cost. \arkot{} is served locally on $4{\times}$A100 80\,GB; baselines are priced at official API rates (June 2026).}
  \label{tab:efficiency}
  \centering
  \footnotesize
  \setlength{\tabcolsep}{4pt}
  \begin{tabular}{@{}l rrrrr@{}}
    \toprule
    Model & Requests & Avg time/item & Total tokens & \$ / item & Total \$ \\
    \midrule
    Gemini-3.5-flash             & $1\,045$ & $240$\,s & $2.03$\,M & \$$0.021$ & \$$8.70$ \\
    DeepSeek-V4-Pro              & $1\,503$ & $582$\,s & $2.83$\,M & \$$0.004$ & \$$1.69$ \\
    GPT-5.2                      & $1\,159$ & $414$\,s & $2.31$\,M & \$$0.018$ & \$$7.35$ \\
    Qwen3.6-max                  & $1\,434$ & $28.2$\,s & $2.46$\,M & \$$0.021$ & \$$8.48$ \\
    Kimi-k2.6                    & $1\,413$ & $28.5$\,s & $1.95$\,M & \$$0.010$ & \$$4.06$ \\
    Claude-4.5-sonnet            & $1\,200$ & $450$\,s & $2.42$\,M & \$$0.044$ & \$$18.14$ \\
    GLM-5.1                      & $1\,618$ & $17.0$\,s & $1.38$\,M & \$$0.005$ & \$$1.98$ \\
    \midrule
    \arkot{} & $971$ & $0.41$\,s & $1.07$\,M & \$$0.0007$ & \$$0.28$ \\
    \bottomrule
  \end{tabular}
\end{table}

\subsection{Qualitative Analysis}
\label{sec:qualitative}

The quantitative metrics above measure geometric agreement but do not fully capture whether the model has learned to \emph{design}---that is, whether it can compose mechanical features into coherent parts for prompts outside its training distribution. \Cref{fig:hero} provides visual evidence along three dimensions.

\paragraph{Spatial reasoning.} Generated parts exhibit correct spatial relationships between features: holes are placed on mounting faces, ribs connect to load-bearing edges, and patterns follow the geometry of the base shape rather than being placed arbitrarily.

\paragraph{Feature composition.} Multi-feature designs combine operations that are individually common (extrude, cut, fillet) into configurations that do not appear in the training data. The model composes these operations in a design-coherent order rather than applying them independently.

\paragraph{Out-of-distribution generalization.} None of the parts in \cref{fig:hero} appears as an identical training example. The model generalizes from learned design patterns to novel geometries, producing parts with correct feature structure for requests that go beyond the training vocabulary.

\paragraph{Remaining failure modes.} The errors that persist concentrate on operations requiring precise coordinate reasoning (revolves, sweeps along complex paths), thin-walled constructions where small numerical errors collapse the geometry, and polar patterns where the model must infer the array axis and count from context. These failures are geometric rather than syntactic---the programs typically execute but produce shapes that deviate from the intended design---and define the frontier for future work on spatial reasoning within parametric construction.

\section{Discussion and Future Work}
\label{sec:conclusion}

\arkot{} reframes text-to-3D generation for engineering use: instead of optimizing for visual plausibility or code executability alone, it targets the structured design state that makes a CAD artifact editable and manufacturable. The experimental results show that a $4$B model trained on design-state-aligned data can match frontier-scale LLMs on geometric quality at substantially lower cost. The qualitative evidence suggests that the model has acquired composable design patterns rather than surface code templates.

Several limitations bound these conclusions. The evaluation relies primarily on geometric metrics (CD, IoU) that can mask missing features; a part can score well while omitting a rib or a bolt pattern. The current pipeline accepts only text and produces single-part designs; multi-part assemblies and non-textual design inputs remain out of scope.

Several directions extend this work naturally. The execution-grounded pipeline can be extended into a self-improving loop: generated programs that pass kernel validation become new training data, compounding the model's coverage over successive rounds. \emph{Multimodal conditioning}---accepting sketches, reference images, or existing CAD geometry alongside text---would bring the design-state approach closer to real engineering workflows, where a designer rarely starts from text alone. Moving from single-part generation to \emph{assembly and iterative editing}---producing multi-part designs with inter-component constraints, and supporting local modifications to an existing design state---would address the full lifecycle of parametric CAD authoring. Developing \emph{feature-level evaluation}---automated checking of whether each requested feature is realized and whether the design remains valid after parameter edits---would close the gap between the design-state formalization and the metrics used to assess it. The scored outputs and evaluation code are slated for public release alongside the final version of this report.

\bibliography{arko_t_refs}

\end{document}